\documentclass[runningheads]{llncs}
\usepackage[T1]{fontenc}
\usepackage{graphicx}
\usepackage{amsmath}
\usepackage{tikz}
\usetikzlibrary{shapes,arrows,positioning}
\usetikzlibrary{fit,backgrounds}

\begin{document}
\title{Multi-Agent Dialectical Refinement \\ for Enhanced Argument Classification}
%
%
\author{Jakub Bąba\orcidID{0009-0009-7000-4887} \and
Jarosław A. Chudziak\orcidID{0000-0003-4534-8652}}
\authorrunning{J. Bąba and J. A. Chudziak}
%
\institute{Faculty of Electronics and Information Technology,\\ Warsaw University of Technology, Poland \\
\email{\{jakub.baba.stud,jaroslaw.chudziak\}@pw.edu.pl}}

\maketitle              
\begin{abstract}
Argument Mining (AM) is a foundational technology for automated writing evaluation, yet traditional supervised approaches rely heavily on expensive, domain-specific fine-tuning. While Large Language Models (LLMs) offer a training-free alternative, they often struggle with structural ambiguity, failing to distinguish between similar components like Claims and Premises. Furthermore, single-agent self-correction mechanisms often suffer from sycophancy, where the model reinforces its own initial errors rather than critically evaluating them. We introduce MAD-ACC (Multi-Agent Debate for Argument Component Classification), a framework that leverages dialectical refinement to resolve classification uncertainty. MAD-ACC utilizes a Proponent-Opponent-Judge model where agents defend conflicting interpretations of ambiguous text, exposing logical nuances that single-agent models miss. Evaluation on the UKP Student Essays corpus demonstrates that MAD-ACC achieves a Macro F1 score of 85.7\%, significantly outperforming single-agent reasoning baselines, without requiring domain-specific training. Additionally, unlike "black-box" classifiers, MAD-ACC's dialectical approach offers a transparent and explainable alternative by generating human-readable debate transcripts that explain the reasoning behind decisions.

\end{abstract}
\section{Introduction}

Argument Mining - the automated extraction and identification of argumentative structures from text - is a crucial field for high-level semantic and reasoning tasks. It enables systems to move beyond surface-level text evaluation, such as grammar or spelling checking, toward deep logic analysis and automated writing evaluation.

Recent developments in Large Language Models (LLMs) have shifted the paradigms used in the field. Current state-of-the-art approaches primarily leverage fine-tuned generative models, achieving high accuracy on tasks such as component classification and relation extraction. However, the reliance on proper tuning of the supervised architecture remains challenging: it is computationally expensive, requires high-quality annotated corpora, and often results in rigid models that struggle to generalize to new domains~\cite{song2025alleviatecatastrophicforgettingllms}. Conversely, training-free LLMs offer a flexible and cost-effective alternative, but they currently fail to bridge the performance gap with supervised baselines. Standard prompting approaches often miss specific details in Argument Mining, resulting in errors, especially between structurally similar components. Moreover, attempts to reduce this via single-agent self-correction mechanisms often result in supporting previously made mistakes instead of correcting them~\cite{chen2025self}. This raises the central question: can LLM-based systems improve their performance in argument component classification by utilizing a multi-agent framework while avoiding the cost of supervised fine-tuning? 

To address this question, we propose the \textbf{MAD-ACC} (Multi-Agent Debate for Argument Component Classification), a framework that formulates argument classification as a structured debate powered by Proponent, Opponent, and Judge agents. Through structured interaction, the model encourages evaluating competing classifications rather than self-refinement. This dynamic enables MAD-ACC to capture and expose logical nuances that are often overlooked by single-pass models. We illustrate this capability through a case study in Section~\ref{sec:illustrative_example}. We show that MAD-ACC reduces the performance gap between inference-only approaches and supervised models, outperforming all evaluated single-pass and reasoning-augmented baselines on the UKP Student Essays corpus~\cite{stab2017parsing}. Moreover, MAD-ACC provides an additional contribution by improving model transparency. Through revealing intermediate arguments in debate and final reasoning behind each decision, the framework offers insight into the decision-making process, addressing a key limitation of black-box classifiers.

\section{Related Work}

The proposed MAD-ACC sits at the intersection of computational argumentation and agentic artificial intelligence. To contextualize our contribution, we survey the literature across the evolution of Argument Mining methodologies, the applications of Large Language Models to these tasks, and the emergence of Multi-Agent Systems for reasoning. We focus there on the paradigm shifts, from feature engineering to deep learning and generative inference. We also review MAS approaches, which motivate our idea of dialectical refinement.

\subsection{Argument Mining Approaches}

Argument Mining (AM)~\cite{lawrence-reed-2019-argument} is a research area within the field of Natural Language Processing, focused on extracting and identifying structured reasoning from unstructured text. The field includes several different subtasks, ranging from boundary identification and relation extraction to the classification of the elements. Foundational work in AM focused on providing annotation schemes and corpora that allowed for structuring and indexing retrieved annotations. To benchmark progress, the community established various domain-specific datasets. Among these, the UKP Argument Annotated Essays corpus~\cite{stab2017parsing} has emerged as a widely adopted standard for analyzing argumentation in educational texts.

Methodologically, this area has experienced a notable shift. Early approaches relied primarily on manual feature engineering, typically combining Support Vector Machines (SVMs) with carefully designed lexical and structural features~\cite{stab2017parsing,habernal-gurevych-2017-argumentation}. As deep learning techniques matured, the state-of-the-art shifted toward neural architectures~\cite{eger-etal-2017-neural,niculae-etal-2017-argument}. Transformer-based models such as BERT and RoBERTa have set new performance benchmarks~\cite{mushtaq2022argument}. Despite their strong accuracy, these supervised approaches remain limited by their dependence on large-scale annotated data, suffering from poor generalization when applied to out-of-domain text.

\subsection{Large Language Models in Argumentation}

The rise of generative models has resulted in another shift, moving from encoder-only architectures to generative Large Language Models (LLMs). This led to an exploration of the training-free capabilities, where models such as GPT-4~\cite{achiam2023gpt} were evaluated on argument mining tasks. Comprehensive evaluations have shown the potential of LLMs~\cite{zhao2023survey,chang2024survey,gorur2024largelanguagemodelsperform}, including the field of argument mining~\cite{pojoni2023argument,al2023performance}. One of the research directions to enhance LLM performance on argument mining was chain-of-thought (CoT) prompting, a promising training-free technique that improves reasoning capabilities on complex tasks~\cite{wei2022chain}. However, the effectiveness of CoT varies significantly on the model size and task characteristics, with recent findings suggesting decreasing results for non-reasoning models~\cite{meincke2025prompting}.

In response to the limitations of prompting, recent state-of-the-art research focused on applying supervised strategies into LLM methods. Recent studies, including Cabessa et al. (2025) demonstrated that fine-tuning of the LLMs achieves superior performance compared to the earlier benchmarks~\cite{cabessa-etal-2025-argument}. Beyond argument mining, fine-tuning has proven to be promising for enhancing reasoning capabilities across various NLP tasks~\cite{pareja2024unveiling,bousselham2024fine}. However, while these fine-tuned approaches currently define the standard, they reintroduce the issue of heavily relying on high-quality annotated data, limiting their usage in low-resource domains and languages, where such annotations are unavailable.

\subsection{LLM-based Multi-Agent Systems}

Multi-Agent Systems (MAS), one of the recently emerging research directions~\cite{tran2025multiagentcollaborationmechanismssurvey}, leverage the concept of structured roles and collaboration to enhance problem solving. By distributing tasks across specialized agents with distinct roles, personas, and specific contexts, MAS frameworks often offer improved problem solving across diverse fields~\cite{kostka2025cognitivesynergyllmbasedmultiagent,zamojska2025gamesagentsplaytransactional}. These systems have demonstrated particular promise in domains such as legal reasoning and Natural Language processing~\cite{gorur2025retrievalargumentationenhancedmultiagent,Sadowski_2025}.

A subset of this field is Multi-Agent Debate (MAD), a concept growing in Argument Mining and NLP. Distributing reasoning among specialized agents engaged in structural discussion enables models to refine through debate and critique~\cite{wu2025llmagentsreallydebate,harbar2025simulating}. Recent research included different MAD frameworks and configurations across diverse NLP tasks~\cite{liu2024groupdebate,estornell2025acccollabactorcriticapproachmultiagent,gou2024criticlargelanguagemodels}. A key recent work~\cite{ku-etal-2025-multi} utilized a debate framework to evaluate implicit premises, outperforming both neural baselines and single-agent LLMs. This effectively showed that agents can achieve better accuracy by discussing and refining their answers based on opposing opinions than by repeated generation.

\section{Methodology}

We propose \textbf{MAD-ACC} (Multi-Agent Debate for Argument Component Classification), a framework designed to resolve ambiguities in the classification task without reliance on annotated training data. In our framework, we leverage dialog to adjudicate competing interpretations of structural relationships within arguments.

\subsection{Task Formulation}

We formalize the \textbf{Argument Component Classification (ACC)} as a sequence labeling task. Let $D = \{t_1, t_2, \dots, t_n\}$ be an argumentative document consisting of $n$ argument components. For each target component $t_i$, let $\mathcal{C}_i$ denote its context window (e.g. whole document or surrounding paragraph).

The objective is to establish a mapping function $\Phi: (t_i, \mathcal{C}_i) \to y$ that assigns the correct label $y \in \mathcal{Y}$, derived from the annotation scheme defined by Stab and Gurevych~\cite{stab2014annotating}:

\begin{equation*}
    \mathcal{Y} = \{ \textsc{MajorClaim}, \textsc{Claim}, \textsc{Premise} \}
\end{equation*}

where classes are defined as follows:
\begin{itemize}
    \item \textbf{\textsc{MajorClaim}:} The root node of the argument structure, represents central thesis of the document.
    \item \textbf{\textsc{Claim}:} An intermediate node that receives support, functions as the topic for evidentiary statements.
    \item \textbf{\textsc{Premise}:} A leaf node that provides support (example, evidence, reason) to Claim or another Premise.
\end{itemize}

\subsection{The MAD-ACC Framework}

We formalize the MAD-ACC framework as a Multi-Agent System (MAS) tuple $\mathcal{S} = \langle \mathcal{A}, \mathcal{P}, \mathcal{T} \rangle$, where:
\begin{itemize}
    \item $\mathcal{A} = \{ \text{Mgr}, \text{Prop}, \text{Opp}, \text{Jud} \}$ is the set of Agents (Manager, Proponent, Opponent, Judge).
    \item $\mathcal{P}$ is the set of agent-specific system Prompts defining their roles.
    \item $\mathcal{T}$ is the shared state (Transcript) of the interaction.
\end{itemize}

\begin{figure}[t!]
\centerline{\includegraphics[width=0.5\textwidth]{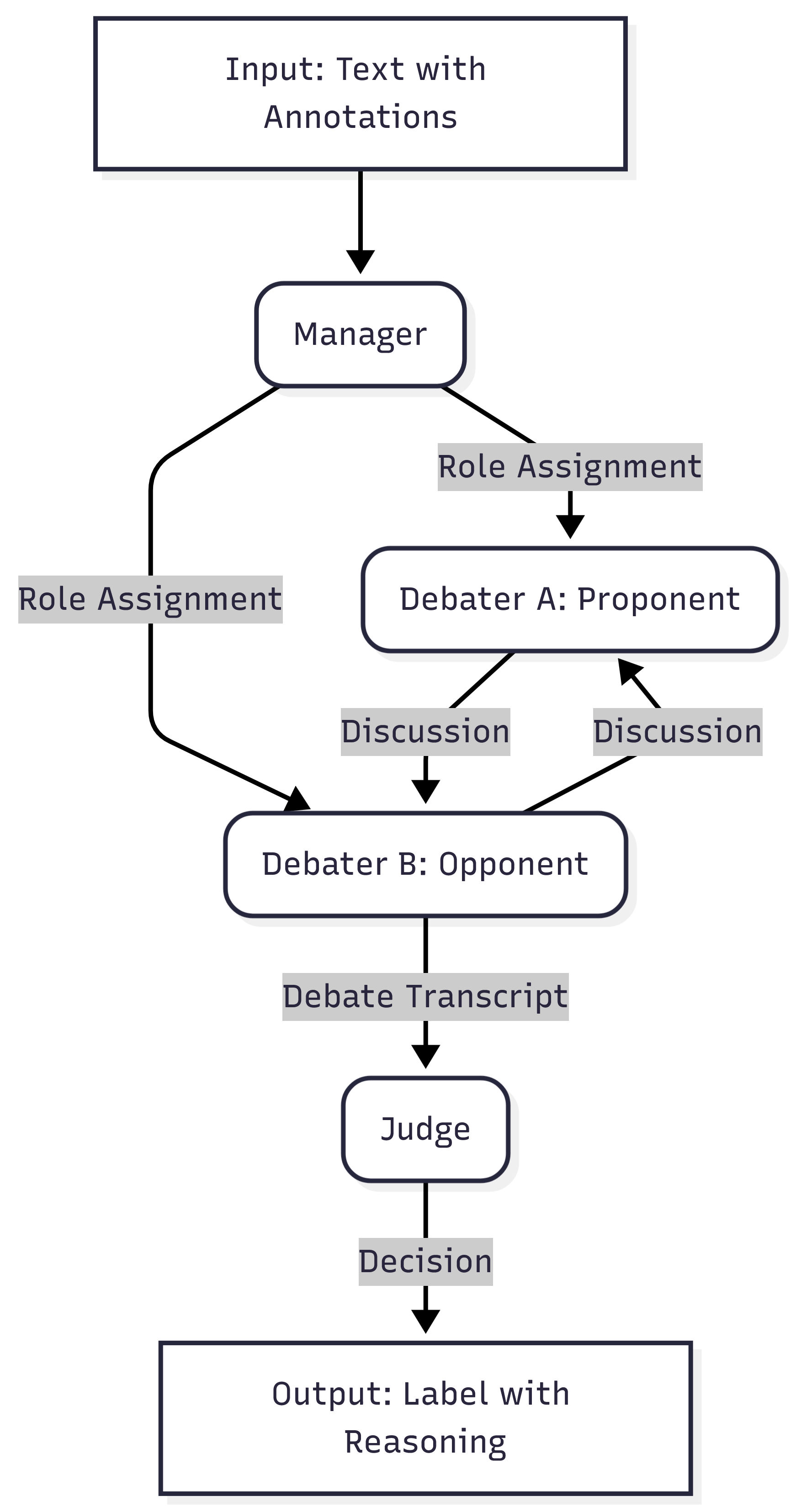}}
\caption{The MAD-ACC system overview.}
\label{fig:system_overview}
\end{figure}

The execution flow of the MAD-ACC framework is illustrated in Figure~\ref{fig:system_overview}. It consists of three phases.

\subsubsection{Probabilistic Initialization}

To properly induce dialectical diversity between debaters, the \textbf{Manager} agent acts as a probabilistic filter. Given input $x$, it estimates the probability distribution over the labels:
\begin{equation*}
    P(y|x) = \text{Mgr}(x) \quad \text{for } y \in \mathcal{Y}
\end{equation*}
Let $y_{top1}, y_{top2} \in \mathcal{Y}$ be the labels with the highest probabilities. To mitigate position and authority biases, system randomly assigns these labels to the \textbf{Proponent} and \textbf{Opponent}, ensuring fairness of the debate for both \textbf{Debaters}. 

\subsubsection{Dialectical Interaction}

The debate is modeled as a sequence of message turns $\mathcal{T} = [m_1, m_2, \dots, m_k]$, where $k$ is total number of turns. At each turn $i$, an active agent $a \in \{\text{Prop}, \text{Opp}\}$ generates a message defending their $y$ label, based on the input $x$ and the conversation history $\mathcal{T}_{<i}$:
\begin{equation*}
    m_i = a(x, y, \mathcal{T}_{<i})
\end{equation*}

The \textbf{Debaters} are instructed to defend their views and to prioritize structural and logical dependencies over isolated semantic assertiveness, explicitly arguing the role of target unit in the context.

\subsubsection{Judge Classification}

The final classification is performed by the \textbf{Judge} agent. The Judge evaluates the evidence provided by both Debaters in the transcript $\mathcal{T}$ against the provided definitions of labels in $\mathcal{Y}$ and proposes a prediction $\hat{y}$:
\begin{equation*}
    \hat{y} = \text{Jud}(x, \mathcal{T}, \mathcal{Y})
\end{equation*}
The Judge resolves ambiguities by verifying which debater provided correct evidence and direction of support, rather than simply aggregating votes.

\subsection{Illustrative Example}\label{sec:illustrative_example}

\begin{figure}

\centering
\begin{tikzpicture}[
    node distance=1.5cm and 0.5cm,
    font=\sffamily\footnotesize,
    box/.style={draw, rounded corners, align=center, fill=white, inner sep=5pt},
    agent/.style={draw, circle, minimum size=0.9cm, fill=gray!15, font=\bfseries},
    speech/.style={draw, cloud, cloud puffs=11, aspect=3, fill=white, align=center, inner sep=1pt, font=\scriptsize},
    arrow/.style={-latex, thick}
]

\node[box, fill=blue!5] (input) {
    \textbf{Input:} "...people often argue that \textbf{\textcolor{blue}{<T>an apartment is more expensive</T>}}. \\
    However, this is only partially true..."
};

\node[agent, below=0.7cm of input] (mgr) {Mgr};
\node[below=0.0cm of mgr, align=left, font=\scriptsize] {
    $P(\text{Premise}) = 0.75$ \\
    $P(\text{Claim}) = 0.20$
};
\draw[arrow] (input) -- (mgr);

\node[below=3.2cm of mgr] (center) {};

\node[agent, left=1.3cm of center] (aga) {Prop};
\node[agent, right=1.3cm of center] (agb) {Opp};

\node[speech, above left=0.1cm and 0.1cm of aga] (speechA) {
    \textbf{Argues Premise:}\\
    "It states a fact.\\
    Supports topic."
};

\node[speech, above right=0.1cm and 0.1cm of agb] (speechB) {
    \textbf{Argues Claim:}\\
    "No, look at 'However'.\\
    It's a Counter-Claim."
};

\draw[arrow, dashed] (mgr) -| (aga) node[pos=0.5, above] {\textit{Assign: Premise}};
\draw[arrow, dashed] (mgr) -| (agb) node[pos=0.5, above] {\textit{Assign: Claim}};

\draw[arrow, bend left=15] (aga) to node[midway, above, font=\scriptsize] {Interaction} (agb);
\draw[arrow, bend left=15] (agb) to node[midway, below, font=\scriptsize] {Interaction} (aga);

\begin{scope}[on background layer]
    \node[draw=gray, dashed, rounded corners, fit=(aga) (agb) (speechA) (speechB), 
    label={[gray, anchor=north east]south east:\textit{Phase 2: Multi-Turn Interaction}}] (debatebox) {};
\end{scope}

\node[agent, below=1.6cm of center] (jud) {Jud};
\draw[arrow] (center |- debatebox.south) -- (jud) node[midway, right, font=\scriptsize] {Transcript};

\node[box, below=0.6cm of jud, fill=green!10] (output) {
    \textbf{Verdict:} Agent B correctly identifies the argument structure. \\
    \textbf{Final Label: CLAIM}
};
\draw[arrow] (jud) -- (output);

\end{tikzpicture}
\caption{Illustrative execution trace.}
\label{fig:illustrative_example}
\end{figure}
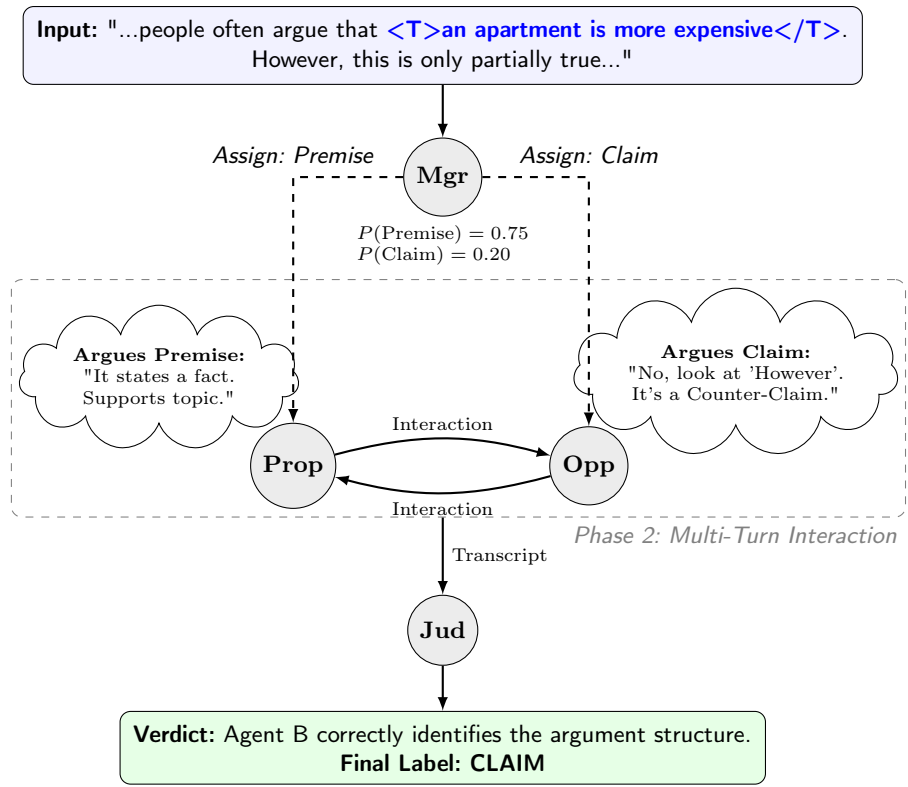

To demonstrate the MAD-ACC capability of resolving ambiguity through contradicting opinions, we present a representative flow in Figure~\ref{fig:illustrative_example}.

The document examines the differences and trade-offs between living in university dormitories and apartments. The target sentence - \textit{"an apartment is more expensive"} - illustrates a case of classic claim ambiguity. Semantically, the statement sounds as a factual observation about accommodation costs. As a result, standard models (and our \textbf{Manager} agent) frequently misclassify this component as a \textit{Premise}, assuming it serves as a supporting evidence rather than as a claim.

However, the debate exposes the true structural role of the sentence in the document. \textbf{Proponent} initially argues for the \textit{Premise} label, interpreting the sentence as a previously stated observation. On the other side of the debate, \textbf{Opponent} identifies the structural function of the argument as a core economic argument in a section of text, directly supporting text's main thesis. The Opponent shows that the subsequent statements refute this statement (e.g. \textit{"this is only partially true"}), and as a result, it functions as a \textit{Claim}.

The \textbf{Judge} agent, leveraging the hierarchical definitions, analyzes the debate transcript and adjudicates in favor of the Opponent. The verdict relies on the direction of support - since the component was recipient of the logic behind the whole paragraph, it hierarchically functions as a \textit{Claim}. This example highlights the core contribution of our work: by forcing agents to debate the structure of the text, defending even unpopular labels, MAD-ACC successfully finds the small details that other classification models miss.

\section{Experiments}

In this section, we present the evaluation process of the MAD-ACC framework. Our primary objective is to check if the proposed dialectical interaction can efficiently increase the performance of the training-free inference. We detail the benchmark dataset and its preparation, the selection of baselines and the specific configuration of multi-agent architecture used to validate our claims regarding accuracy and interpretability. 

\subsection{Dataset}

For our analysis, we used the \textbf{UKP Argument Annotated Essays v2}~\cite{stab2017parsing} corpus, a dataset containing 402 essays and 6089 statements. To ensure strict comparability with state-of-the-art supervised solutions, experiments were conducted based on the exact test split (80 essays with 1266 argument statements) established in prior literature~\cite{cabessa-etal-2025-argument}. No portion of the training split was used for prompt calibration or manual tuning.

Prior to processing by the MAD-ACC framework, the corpus with the annotations was formatted to enable easier LLM agent comprehension. For each instance, the full essay was provided, with argument components delimited by the tags. The target component was marked as \texttt{<TARGET>...</TARGET>}, while surrounding components were masked as generic \texttt{<ARG>...</ARG>}, without leaking the ground truth labels.

\subsection{Baselines}

We evaluate performance using Macro F1, Weighted F1 and class-specific F1 scores for MajorClaim, Claim and Premise types. To validate the effectiveness of the multi-agent framework, we compare MAD-ACC against three different, single-agent baselines:
\begin{enumerate}
    \item \textbf{Vanilla:} Represents standard usage of the LLMs. It utilizes the same model as the Manager agent (\textit{Gemini 2.5 Flash}), with the direct classification prompt.
    \item \textbf{Chain-of-Thought (CoT):} Utilizes standard Chain-of-Thought reasoning prompting with \textit{Gemini 2.5 Flash} to assess if internal reasoning is sufficient.
    \item \textbf{Smart Reasoning:} Uses the more capable \textit{Gemini 2.5 Pro} model with built-in reasoning and the exact same system definitions and rules as ones given to a Judge agent. It is designed to simulate Judge's decision making process without the benefit of the debate content.
\end{enumerate}

Additionally, we contextualize our results with the state-of-the-art supervised approaches, specifically fine-tuned LLMs~\cite{cabessa-etal-2025-argument}. While these methods currently define the upper bound baseline, we highlight that they act as "black-box" solutions with limited explainability, whereas our framework prioritizes transparency and reasoning used behind decisions.

\subsection{Experimental Setup}

For this study, we set the debate length to 2 rounds (four total turns), allowing each agent to present its initial argument and respond to the opponent's counterargument. This configuration reflects a 
trade-off between argumentative depth and efficiency: preliminary experimentation and manual inspection indicated that a single round often fails to expose structural disagreements, while longer debates tend to introduce repetitiveness without yielding additional classification benefits. 

To mitigate the position bias from the Judge agent, we employed a randomized stance assignment strategy. Proponent and Opponent agents are randomly assigned to defend first and second most probable label, ensuring that the order of probabilities will not affect the final judgment. While the framework supports a confidence-based skip threshold, we treated all samples with the debate to rigorously evaluate system's ability to resolve ambiguities in the corpus.

We used the \textbf{Gemini 2.5} family of models. The Manager and the Judge agents used \textit{Gemini 2.5 Flash} and \textit{Gemini 2.5 Pro} respectively; the Manager was designed to quickly filter the least probable label, while the Judge required higher capacity to process debate context. While the MAD-ACC framework is model-agnostic, we prioritized establishing a strong baseline with Gemini and leave the comparative analysis of other models, including open-source alternatives, for future work. For both agents, the temperature was set to \textbf{0.0} to ensure deterministic outputs and consistent scoring. The Debaters used \textit{Gemini 2.5 Flash} model, with a temperature of \textbf{0.7}, selected to ensure creativity through diverse reasoning paths during label defense.

\section{Results and Discussion}

In this section, we present the empirical evaluation of the MAD-ACC framework on the UKP Student Essays corpus. We analyze the MAD-ACC performance against selected single-agent baselines and contextualize them with a supervised approach. Subsequently, we conduct a qualitative analysis of selected examples based on the debate transcripts, to show the power of our system in correcting logical errors through adversial reasoning, highlighting the interpretability benefits of our framework. 

\subsection{Performance Analysis}

\begin{table}[b]
\caption{Comparison of classification performance on UKP Student Essays.}
\begin{center}
\begin{tabular}{|l|c|c|c|c|c|}
\hline
\textbf{Method} & \multicolumn{2}{|c|}{\textbf{Overall Performance}} & \multicolumn{3}{|c|}{\textbf{Class-wise F1-score}} \\
\cline{2-6} 
\textbf{} & \textbf{\textit{Macro F1}} & \textbf{\textit{W-F1}} & \textbf{\textit{MC}} & \textbf{\textit{Claim}} & \textbf{\textit{Premise}} \\
\hline
\multicolumn{6}{|l|}{\textit{Inference-Only Baselines}} \\
\hline
Baseline A (Vanilla) & 78.5 & 80.8 & 90.6 & 57.0 & 88.0 \\
Baseline B (Chain-of-Thought) & 79.2 & 81.2 & 91.4 & 58.5 & 87.8 \\
Baseline C (Smart Reasoning) & 84.9 & 86.1 & \textbf{92.2} & 72.5 & 90.1 \\
\textbf{MAD-ACC (Ours)} & \textbf{85.7} & \textbf{87.0} & 92.0 & \textbf{74.5} & \textbf{90.7} \\
\hline
\multicolumn{6}{|l|}{\textit{Supervised Reference}} \\
\hline
Cabessa et al. (2025) & \textbf{89.5} & - & - & - & - \\
\hline
\end{tabular}
\label{tab:main_results}
\end{center}
\end{table}

Table~\ref{tab:main_results} summarizes the argument classification results on the UKP test set. The results show a clear performance hierarchy. Context-free, general-purpose inference baselines (Baselines A and B) achieve Macro F1 scores of 78.5\% and 79.2\%, respectively. Single-agent reasoning (Baseline C) achieves a Macro F1 score of 84.9\%, and MAD-ACC achieves the highest training-free performance of 85.7\%.

Chain-of-Thought achieves slightly better performance than vanilla prompting, however both of the baselines hit the ceiling of approximately 80\%. Moreover, both of them mostly struggle with Claim components (Claim F1 scores of 57.0\% and 58.5\%). This supports the theory of incorrect reliance on surface-level semantics and often incorrectly connecting components "sounding like opinions" with Claims.

There is a major improvement between the first two baselines and the solutions equipped with base rules. Baseline C, equipped with reasoning and such knowledge, improves consistently, particularly in Macro F1, and more importantly, Claim F1, which is up around \textbf{+14 percentage points}. It suggests that moving the focus to resolving semantic ambiguities based on direction of support substantially improves performance.

Utilizing dialectical refinement in the pipeline resulted in MAD-ACC achieving the best results, beating Baseline C by ~0.8\% in Macro F1. Moreover, the MAD-ACC outperformed strong single-agent baseline by another Claim F1 \textbf{+2\%}, while keeping F1 scores for Premises and Major Claims stable. This means that it MAD-ACC isn't just moving classifications from Premises to Claims, but actively using debate to differentiate components more effectively.

\subsection{Comparison with State-of-the-Art}

Table~\ref{tab:main_results} contextualizes our results against the supervised state-of-the-art~\cite{cabessa-etal-2025-argument}. While fine-tuned LLMs currently define the upper bound performance of 89.5\% Macro F1 score, MAD-ACC reduces this gap with a competitive 85.7\% without requiring any training or parameter updates.

This result highlights a trade-off between \textit{Performance} and \textit{Data Efficiency}. Their SOTA model achieved better results, but fine-tuning relied on approximately 80\% of the corpus, whereas MAD-ACC operates in a training-free setting. Ultimately, fine-tuning remains optimal for the cases where the cost of tuning is acceptable and annotated data is available. On the other side, the MAD-ACC framework presents a compelling alternative for low-resource domains and cases where annotating large amounts of documents is impossible. We note that while MAD-ACC eliminates training costs, the multi-agent debate increases costs from token consumption compared to single-pass prompting. However, for low-resource domains, this trade-off is often preferable to data annotation costs.

\subsection{Qualitative Analysis: Case Studies}\label{sec:qualitative_analysis}

To investigate the source of MAD-ACC performance gain, we analyzed exemplary instances where the single-agent baseline (Baseline C) failed, but MAD-ACC labeled the component correctly by taking advantage of reasoning from the debate transcript. 

\begin{table}[ht]
\centering
\caption{Qualitative comparison of Baseline C vs. MAD-ACC. The dialectical transcript allows the Judge to resolve structural ambiguity where the single-agent fails.}
\label{tab:case_study}
\begin{tabular}{|p{0.15\textwidth}|p{0.35\textwidth}|p{0.45\textwidth}|}
\hline
\textbf{Case Type} & \textbf{Text \& Context} & \textbf{Dialectical Resolution (MAD-ACC)} \\
\hline
\textbf{Case 1:} \newline Topic \newline Sentence \newline (Essay 335) & 
\textit{Target:} "connecting people by email is easy and fast" \newline
\textit{\textbf{Predictions:}} \newline
\textit{Baseline:} \textcolor{red}{\textbf{Premise}} \newline
\textit{MAD-ACC:} \textcolor{green}{\textbf{Claim}} & 
\textit{\textbf{Agent B Argues (simplified):}} The target sentence is a Claim, as it is presented as argument supporting another idea that IT has benefits, which is a crucial component of the essay's overall thesis. At the same time, it acts as the core ideas of the paragraph, one of direct benefits. \newline
\textit{\textbf{Verdict:}} Judge accepts that the target is one of the main arguments presented in the paragraph and functions as a Claim. \\
\hline
\textbf{Case 2:} \newline Hierarchy Resolution \newline (Essay 169) & 
\textit{Text:} "Only by developing students, can we have a better academic field" \newline
\textit{\textbf{Predictions:}} \newline
\textit{Baseline:} \textcolor{red}{\textbf{MajorClaim}} \newline
\textit{MAD-ACC:} \textcolor{green}{\textbf{Claim}} & 
\textbf{\textit{Agent A Argues (simplified):}} The target is a Main Argument and a Claim that supports the essay's overarching thesis, not the thesis itself. It provides the abstract principle explaining why professors should prioritize teaching. It connects directly to the MajorClaim by justifying it, functioning as a support rather than the root node. \newline
\textbf{\textit{Verdict:}} Judge recognizes the target as a Claim as a high-level argument supporting the Thesis. \\
\hline
\end{tabular}
\end{table}

\subsubsection{Resolving Topic Sentence Ambiguity (Case 1)} 
Single-pass models often confuse the main argument with the evidence supporting it, especially when the argument is descriptive. In Essay 335, the baseline incorrectly classified the component "connecting people by email is easy and fast" as a Premise, underestimating its role in the text. However, the debate transcript shows that Agent B correctly identified it as one of the main arguments in the text, directly supporting the main thesis of the essay (\textit{"IT discoveries are likely to have more disadvantages than benefits and people should know how to use their developments properly"}), and as a result, it was correctly classified as a Claim. 

\subsubsection{Hierarchical Distinction (Case 2)}
These models also sometimes struggle with abstraction and hierarchy. In Essay 169, the target sentence "Only by developing students, can we have a better academic field" was misclassified by a baseline as a MajorClaim. The debate process correctly found and analyzed the dependency chain, realizing that while the sentence was abstract, it was a pillar for the main thesis of the text (\textit{"professors should spend more time on preparing courses than research"}), and as a result, it was a Claim.

\subsubsection{}
These examples demonstrate that the dialectical process forces the system to thoroughly evaluate the function of each component and decide whether it supports a neighbor (Premise), the thesis (Claim) or acts as a main idea - and as a result, it supports resolving structural ambiguity present in the baselines. 

\section{Future Work}

The results presented in this work suggest several promising directions for future research.

Firstly, the framework's cross-domain generalization capabilities should be investigated. Since MAD-ACC does not rely on domain-specific annotated data, it is a strong candidate for application in other domains, especially low-resource ones, including legal, political, or biomedical text mining. Future studies could evaluate the usage of such a dialectical framework, especially in the role of an assistant suggesting initial annotations, with the powerful reasoning helping in validating these labels.

A second area of study is the extension of the approach beyond component classification. The framework could be assessed on other tasks, such as Argument Relation Identification and Classification (ARI/ARC). By configuring agents to debate the existence and types of links (Support/Attack), subsequent research could move toward full argument structure parsing.

Finally, a critical direction for such agentic systems would be to integrate them into real-world educational technologies. Future work could deploy MAD-ACC within intelligent educational systems, where the framework's explainability and interpretability could be measured through pedagogical impact and user studies. Such studies could verify the value of the AI-generated dialectical annotations in the process of improving argumentative skills.

\section{Conclusion}

In this work, we presented the \textbf{MAD-ACC}, a multi-agent framework utilizing dialectical refinement to improve performance on the Argument Component Classification task without relying on expensive fine-tuning on high-quality annotated data. By replacing static classification with a multi-agent debate of contradicting opinions, we addressed the limitations of single-agent LLMs, specifically their tendency to mismatch the structural function of the argument based on a semantic assertiveness.

Our experiments on the UKP Student Essays corpus demonstrate that MAD-ACC achieves a Macro F1 score of 85.7\%, outperforming all the baselines without task-specific training. Notably, the framework effectively resolves the "Claim vs. Premise" ambiguity, providing substantial improvement in the Claim F1 score. As demonstrated by the qualitative analysis in Section~\ref{sec:qualitative_analysis}, the debate mechanism successfully corrects errors where single agents misclassify topic sentences or hierarchy of the documents, based on the direction of support. While state-of-the-art still holds a performance advantage (89.5\%), our methodology provides a competitive, data-efficient alternative for low-resource domains.

Beyond quantitative performance, MAD-ACC adds a level of explainability that is largely absent in traditional classifiers. The generated debate transcripts provide a transparent thought process behind each decision, shifting the system from a black-box model to a tool capable of justifying its conclusions to users.

\subsubsection{\ackname} The work reported in this paper was supported by the Polish National Science Centre under grant 2024/06/Y/HS1/00197.
 
%
%
%
\bibliographystyle{splncs04}
\bibliography{bibliography}

\end{document}